\documentclass{article}
\usepackage{spconf,amsmath,graphicx}
\usepackage{subfig}
\usepackage{array,booktabs}
\usepackage[flushleft]{threeparttable}
\usepackage{arydshln}
\usepackage[font=small,labelfont=bf]{caption}
\usepackage{multirow,multicol}
\usepackage{caption}
\usepackage{xcolor}
\usepackage{soul}
\usepackage{enumitem}

\usepackage[hyperfootnotes=false]{hyperref}

\usepackage{caption}
\makeatletter
\def\@name{ \emph{Niloufar Alipour Talemi\,\orcidlink{0009-0000-6881-3671}, Hossein Kashiani, Sahar Rahimi Malakshan},  \\ \emph{Mohammad Saeed Ebrahimi Saadabadi, Nima Najafzadeh, Mohammad Akyash, Nasser M. Nasrabadi}}
\makeatother

\usepackage{orcidlink}
\title{AAFACE: Attribute-aware Attentional Network for Face Recognition}
\address{West Virginia University}
\begin{document}
\maketitle

\begin{abstract}

In this paper, we present a new multi-branch neural network that simultaneously performs soft biometric (SB) prediction as an auxiliary modality and face recognition (FR) as the main task. Our proposed network named AAFace utilizes SB attributes to enhance the discriminative ability of FR representation. To achieve this goal, we propose an attribute-aware attentional integration (AAI) module to perform weighted integration of FR with SB feature maps. Our proposed AAI module is not only fully context-aware but also capable of learning complex relationships between input features by means of the sequential multi-scale channel and spatial sub-modules. Experimental results verify the superiority of our proposed network compared with the state-of-the-art (SoTA) SB prediction and FR methods.

\end{abstract}
\section{Introduction}
\label{sec:intro}

FR has been one of the most popular research areas in computer vision due to its vital role in safety and security applications. In the last decade, the introduction of convolutional neural networks (CNNs), and different margin-based loss functions \cite{kim2022adaface, arcface} considerably improved the performance of FR. However, the SoTA FR methods experience severe performance degradation in unconstrained scenarios due to several factors such as variances in the head pose, inherent sensor noise, and illumination conditions.

There are several multi-task learning frameworks in biometrics that use shared parameters of CNN and build synergy among the highly related tasks to boost their individual performances \cite{ranjan2017all, ranjan2017hyperface, liu2015, moon, zhuang2018multi, mao2020deep}. For instance, \cite{ranjan2017all, wang2017multi} enjoyed the advantage of a multi-task learning structure to perform FR in addition to facial attributes prediction and some other face-related tasks. However, they do not directly employ attribute information to enhance FR performance while intrinsically humans analyze facial attributes to recognize identities. Even police usually ask witnesses about gender, the shape of the nose, and many other identity-related attributes to reconstruct convicts' faces.

In this work, we utilize SB information as an auxiliary modality to boost the discriminative ability of our model for FR. To this end, we rely on the facial attributes of a subject which stay the same for different images of the same identity. For instance, the gender and shape of the eyes remain the same in different situations such as various illuminations or poses while some attributes like the color of hair may vary in different images of the same person. Thus, as an auxiliary modality, in this work, five SB attributes are considered which are gender, big nose, chubby, narrow eyes, and bald. 

Our model employs an attentional feature-level integration strategy to fuse feature representations of SB and FR. Most existing feature integration methods are partially context-aware and incapable of capturing the inconsistency in the semantic or scale level of the input feature representations \cite{a1, hu2018squeeze}. Also, some integration studies such as \cite{dai2021attentional}
emphasize informative features only across the channel dimension. To address these issues, we propose a fully context-aware integration module that utilizes both input features to compute the attention weights along the channel and also spatial dimensions. Moreover, unlike integration studies that yield a scalar fusion weight \cite{srivastava2015training}, our proposed AAI module produces an integration weight that matches the size of the input feature maps. This would lead to the generalization improvement of our model to identify relevant features across the input feature maps. To the best of our knowledge, the proposed AAFace method is the first work that employs SB attributes through an attention mechanism to enrich the FR feature representations.  In summary, the contributions of this paper include:

\begin{figure*}[t]
    \centering 
    \includegraphics[scale=1.1]{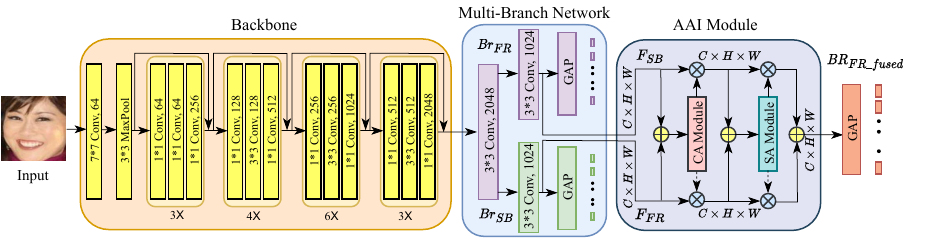}
    \caption{Proposed attribute-aware attentional network for face recognition.}
    \label{fig:Proposed Network}%
\end{figure*}

\vspace{-\topsep}
\begin{enumerate}[align=left,leftmargin=*]
\item We propose a multi-branch neural network that simultaneously performs SB prediction and FR in order to enhance the performance of the FR. To effectively leverage SB information for FR, we adopt a feature-level integration strategy through our AAI module.
\vspace{-\topsep}
\item Our context-aware AAI module employs novel multi-scale attention sub-modules to highlight informative features through both channel and spatial dimensions. Our evaluations exhibit the advantage of our proposed AAI module over other integration methods.
\vspace{-\topsep}
\item Extensive experiments prove that utilizing SB information through the AAI module boosts the performance of FR. Experimental evaluations also demonstrate that our proposed AAFace method outperforms the SoTA SB prediction and FR methods. 
\vspace{-\topsep}
\end{enumerate}

\section{Proposed Method}
\label{sec:format}

 As shown in Fig. \ref{fig:Proposed Network}, the proposed architecture consists of two branches ($Br_{FR}$ and $Br_{SB}$) with the shared backbone which contains convolutional layers of ResNet-50. $Br_{FR}$ is dedicated to perform FR and $Br_{SB}$ is designed to predict SB attributes. Moreover, to enrich the embedding of the FR, an attention mechanism is used to integrate the SB data as auxiliary information with the FR feature maps.

\subsection{Multi-Branch Network}
 For FR, as shown in the $Br_{FR}$ branch of Fig. \ref{fig:Proposed Network}, two convolutional layers are integrated into the backbone. To reduce the dimensions of the feature map, global average pooling (GAP) is applied to the output of the convolutional layers. The output feature representation is then fed to the softmax layer since this branch is intrinsically considered for face identification. In the case of the face verification task, we use the output of the GAP layer as the feature representation of the input data. The cosine distance between a pair of feature representations determines whether they belong to the same identity or not.

The architecture of the $Br_{SB}$ branch which is dedicated to SB prediction is similar to the $Br_{FR}$. As the $Br_{FR}$ branch contains the fine-grained face information that can be leveraged for other face-related tasks, we keep the first convolutional layer of this branch to accurately predict the SB attributes as well. We train binary classifiers to predict different facial attributes, each with its cross-entropy loss. The total classification loss for the $Br_{SB}$ is given by:
\begin{equation} \label{eq1}
L_{SB} = \sum_{i=1}^{n}{\lambda_{a_{i}}L_{a_{i}}},
\end{equation} 
\noindent where each $L_{a_{i}}$ represents a loss for each individual attribute and $\lambda_{a_{i}}$
is the loss-weight corresponding to the attribute $a_{i}$. Also, $n$ denotes the number of SB attributes in $Br_{SB}$. For each attribute, $L_{a_{i}}$ is computed as:
\begin{equation} \label{eq2}
L_{a_{i}} = - \left( a_{i} \log(p_{a_{i}}) + \left(1-a_{i}\right) \log{\left(1-p_{a_{i}}\right)} \right),
\end{equation} 
\noindent where $p_{a_{i}}$ is the probability that the network computes for $a_{i}$.
\subsection{Attentional Integration of SB Attributes with the FR Features}
\begin{figure}[t]
    \centering
    \includegraphics[scale=1.0]{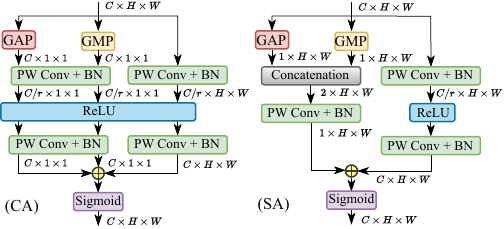}
    \caption{Channel attention (CA) and spatial attention (SA) modules.}
    \label{fig:SA_CA}
\end{figure}

As depicted in Fig. \ref{fig:SA_CA}, the proposed AAI module has two sequential sub-modules which are channel and spatial attention, respectively. Given two feature maps, $F_{FR}$ and $F_{SB}$, the channel-based integration weight, $M_{c}$, is computed from the multi-scale channel sub-module and then this integration weight will be multiplied by $F_{FR}$ feature (i.e., $F_{FR} \times M_{c}$). However, when it comes to the other feature map,  $F_{SB}$, the complementary value of the integration weight will be multiplied by the $F_{SB}$ feature (i.e., $F_{SB} \times (1-M_{c})$). Then, the channel-based weighted averaging between $F_{FR}$ and $F_{SB}$ will be given as input to the multi-scale spatial sub-module. Similar to the channel sub-module, spatial-based weighted averaging will be computed between $F_{FR} \times M_{s}$ and $F_{SB} \times (1-M_{s})$. Therefore, the final fused feature can be formulated as:
\begin{equation} \label{eq3}
F_{fused}=	M_{s} \left( M_{c} \otimes F_{FR} \right) + \left( 1-M_{s} \right) \Big(\big( 1-M_{c} \big) \otimes F_{SB} \Big), 
\end{equation} 

\noindent where $F_{fused}$ is the fused feature, and $\otimes$ denotes the element-wise multiplication.

\noindent\textbf{Channel sub-module}. Considering each channel of a feature map as a feature detector, channel attention focuses on ‘what’ is meaningful given an input image. As shown in Fig. \ref{fig:SA_CA} (CA), we propose a multi-scale channel attention which captures both local and global contexts. To effectively compute the local context, the spatial dimension of the input feature map is aggregated using both GAP and global max pooling (GMP) operations. Then, the two generated local context descriptors are forwarded to a shared bottleneck structure while the global context descriptor is sent to a separate bottleneck structure. As point-wise (PW) convolution only exploits channel interactions for each spatial position, in this work, we utilize it as the local channel context aggregator. It should be noted that broadcasting addition is utilized to merge the local and global contexts. Finally, due to the sigmoid function, $M_{c}$ consists of real numbers between 0 and 1, which enables the network to do weighted averaging between feature representations.

\noindent\textbf{Spatial sub-module}. In comparison with channel attention, spatial attention concentrates on ‘where’ is an informative part, which is complementary to channel attention. Thus, as it is illustrated in Fig. \ref{fig:SA_CA} (SA), both GAP and GMP operations are applied along the channel axis and then their outputs are concatenated to generate an efficient local feature descriptor. Moreover, similar to the channel sub-module, a convolutional bottleneck structure is applied to the global context descriptor to encode where to emphasize or suppress. It is worth noting that both channel and spatial-based integration weights have the same shape as the input features so that the proposed module can preserve and highlight the subtle details in the low-level features.
\subsection{Joint SB Attributes Prediction and the Integrated FR}
\label{ssec:joint}
After feature integration, we utilize the fused feature maps to improve the performance of FR. Therefore, an average pooling followed by a softmax layer is applied to the enriched feature map. Also, since SB prediction is relevant to FR, we train both branches, $Br_{SB}$ and $Br_{FR\_fused}$, simultaneously. Training the two branches jointly, helps the features to gain a better understanding of facial characteristics, which leads to improvements in the performance of individual tasks. As such, the classification loss from both the $Br_{FR\_fused}$ and $Br_{SB}$ branches are backpropagated through the network. Therefore, the total loss is as follows:
\begin{equation} \label{eq4}
Loss_{total}=	\lambda_{FR} L_{FR} + L_{SB} = \lambda_{FR} L_{FR} + \sum_{i=1}^{n}{\lambda_{a_{i}}L_{a_{i}}},
\end{equation} 
\noindent where $L_{FR}$ and $ L_{SB}$ denote the loss from the $Br_{FR\_fused}$ and $Br_{SB}$ branches, respectively. Similar to $\lambda_{a_{i}}$, the weight parameter $\lambda_{FR}$ is determined empirically by considering the importance of our main task which is FR. It should be noted both our $Br_{FR}$ and $Br_{FR\_fused}$ can be trained by any classification loss functions. In this work, we have used one of the recent SoTA margin-based loss functions named AdaFace \cite{kim2022adaface} for training FR branches.

\section{Experiments}
\label{sec:typestyle}
\subsection{Datasets}
We separately train our model on two datasets in order to conduct a fair comparison with other methods. CelebA \cite{liu2015} is a large-scale face attributes dataset with 202,559 face images which covers large variations in pose, background, and illumination. In addition, to demonstrate the generalization of our proposed AAFace method under other training settings, we also employ a portion of WebFace12M \cite{webface260m}, which includes more than 5M face images. Since only the gender facial attribute is provided in this training dataset \cite{huang2021age}, one attribute is used as the auxiliary modality in this case. For evaluation, in addition to the test set of the CelebA dataset, we have also used the LFW \cite{LFWTech}, CFP-FP \cite{CFPFP}, CPLFW \cite{CPLFWTech}, and AgeDB \cite{agedb} datasets which are the most popular benchmarks for FR. Also, we include IJB-B\cite{whitelam2017iarpa} and IJB-C \cite{maze2018iarpa} in our experiments as they are among the most challenging datasets to evaluate unconstrained FR.

\subsection{Implementation and training details}
\label{training}
As for FR, it is important to train our model by means of a large-scale training dataset, in this work, our backbone is weighted with a pre-trained ResNet-50 on the VGGFace2 dataset \cite{cao2018vggface2}. The training process of the proposed method includes three steps. Firstly, we train $Br_{FR}$ when all the layers of the backbone are frozen. Then, the SB branch (named $Br_{SB}$) is separately trained while the backbone and the first convolutional layer of $Br_{FR}$ are kept frozen. Finally, having integrated the feature maps of $Br_{FR}$ and $Br_{SB}$, we jointly train both $Br_{FR\_fused}$ and $Br_{SB}$ branches to efficiently refine feature maps. 
The weight parameters of the total loss function are chosen by considering FR as our main task. We set $\lambda_{FR}$ = 3 , $\lambda_{Male} = \lambda_{Bald} = 1$, and all the rest of the weight parameters equal to $0.5$. The model is trained for $25$ epochs using the stochastic gradient descent, the initial learning rate is set at $0.01$, and the scheduling step at $4$, $10$, and $17$ epochs.
\subsection{Results and Analysis}
\textbf{Soft Biometric Prediction}. Table \ref{Tab:SB} shows the performance of the proposed SB predictor in comparison with the SoTA methods on the CelebA dataset. Similar to the SoTA methods, we have followed the same protocol, and the results of the other methods are directly reported from the original papers. Table \ref{Tab:SB} indicates that the performance of the SB predictor gains considerable improvement when the network is trained for both FR and SB prediction jointly. Furthermore, it shows that our proposed multi-head network outperforms most of the current existing SB prediction methods.

\begin{table}[h!]
\small
\setlength\tabcolsep{3.71pt}
\caption{Reports classification comparison in terms of accuracy (\%) between the proposed SB predictor and the SoTA methods on the CelebA dataset.} 
\centering
\begin{tabular}{l c c c c c} 
\hline \hline
\multirow{2}{*}{Methods} & Bald & Big Nose & Chubby & Male & Narrow Eye \\
{} & (B) & (BN) & (CH) & (M) & (NE) \\
\hline
Z. Liu \cite{liu2015} & 98.00 & 78.00 & 91.00 & 98.00 & 81.00\\
Moon \cite{moon} & 98.77 & 84.00 & 95.44 & 98.10 & 86.52\\
HyperFace \cite{ranjan2017hyperface} & - & - & - & 97.00 & -\\
R. Ranjan \cite{ranjan2017all} & - & - & - & 99.00 & -\\
MCFA \cite{zhuang2018multi} & 99.00 & 84.00 & 96.00 & 98.00 & 87.00\\
L. Mao \cite{mao2020deep} & 99.03 & 84.78 & 95.86 & 98.29 & 87.73\\
Ours & 98.14 & 83.27 & 95.48 & 98.74 & 85.26\\
Ours (jointly) & \textbf{99.10} & \textbf{84.84} & \textbf{96.09} & \textbf{99.16} & 87.56\\
\hline \hline
\end{tabular}
 \label{Tab:SB}
\end{table}

\noindent\textbf{Face Recognition}.
To evaluate our FR model as a verifier, we randomly selected $10,000$ pairs from the CelebA dataset whose identities are not included in the training set. Regarding section \ref{training}, in the first step of the training process, we train $Br_{FR}$ separately. Thus, we can consider this branch as a baseline to better clarify the effective role of integrating SB in FR. Experimental results included in Table \ref{Tab:base} validate that our proposed model improves FR performance by leveraging identity facial attributes. We have also employed different integration strategies instead of our proposed AAI module to demonstrate its advantage over them. Experiments show that by replacing the AAI module with simple operations like addition or concatenation, the performance of the network even underperforms the baseline in some of the false acceptance rates (FARs). Moreover, results prove that among recent feature integration methods, the AAI module builds the most effective integration for improving the performance of FR.  We have also considered different aggregation scales in our attention module to find out the best scale for aggregating the channels. As shown in Table \ref{Tab:base}, the proposed model achieves the best performance with setting $r = 8$. Further, more experiments are implemented to explore the effect of the number of attributes utilized for FR. Regarding the last three rows of Table \ref{Tab:base}, utilizing more identity facial attributes helps the network to perform more accurately.

As mentioned before, for all the experiments in Table \ref{Tab:SB} and Table \ref{Tab:base}, the CelebA dataset is used as the training set. However, to gain a better insight into the advantage of utilizing attributes for FR, we have also used a portion of WebFace12M \cite{webface260m}, which includes more than 5M face images as a training set. We have followed all the settings in AdaFace including the same backbone (resnet101), loss function, and training data. Table \ref{Tab:FRR}, verifies that despite the availability of only one attribute for this training set, our proposed method outperforms AdaFace in most of the benchmarks. It is worth noting that the reason some other methods have slightly better performance than our model in some benchmarks is due to their different FR loss functions. Regarding the flexibility of our model to use any loss function for training the FR branches, we believe that incorporating other loss functions into our model can also improve the performance of those other methods. 
\begin{table}[t]
\small
\setlength\tabcolsep{4.8pt}
\caption{Performance comparison between the proposed method (AAFace), the baseline, and other SoTA feature integration methods. Also, different settings for the proposed AAFace method are ablated. Results are based on TAR$@$FAR, in which TAR and FAR stand for True Acceptance Rate, and False Acceptance Rate, respectively.} 
\centering
\begin{tabular}{l c c c c c c} 
\hline \hline
\addlinespace[1mm]
Methods & $10^{-5}$ & $10^{-4}$ & $10^{-3}$ & $10^{-2}$ & $10^{-1}$\\ [1ex]
\hline
Baseline (without SB) & 87.98 & 89.47 & 91.23 & 92.64 & 93.92\\
Concatenation & 87.62 & 89.03 & 91.19 & 92.95 & 94.14\\
Addition & 87.33 & 89.41 & 90.89 & 92.82 & 94.03\\
SENET \cite{hu2018squeeze} & 89.73 & 90.91 & 92.77 & 94.27 & 95.53\\
AFF \cite{dai2021attentional} & 89.85 & 91.60 & 92.86 & 94.39 & 95.72\\
\cdashline{0-6}
AAFace ($r=4$) & 89.87 & 91.63 & 92.84 & 94.32 & 95.67\\
AAFace ($r=16$) & 90.12 & 92.02 & 92.93 & 94.50 & 95.71\\
AAFace ($r=8$) & \textbf{90.21} & \textbf{92.11} & \textbf{93.09} & \textbf{94.52} & \textbf{95.81}\\
\cdashline{0-6}
AAFace (M) & 88.86 & 90.51 & 91.68 & 93.71 & 94.93\\
AAFace (M \& B) & 89.13 & 90.66 & 92.01 & 93.91 & 94.99\\
AAFace (M \& B \& Ch) & 89.27 & 90.89 & 92.34 & 94.20 & 95.53\\
\hline \hline
\end{tabular}
  \label{Tab:base}
\end{table}

\begin{table}[t]
\small
\setlength\tabcolsep{2.04 pt}
\caption{Performance comparison of our proposed method (AAFace) with recent SoTA FR methods. TAR is reported at FAR = $0.01\%$.} 
\centering
\begin{tabular}{l | c c c c | c c} 
\hline \hline
\multirow{2}{*}{Methods} & \multicolumn{4}{c|}{Verification Accuracy} & \multicolumn{2}{c}{TAR}  \\ \cline { 2 - 7 } & LFW & CFP-FP & CPLFW & AgeDB & IJB-B& IJB-C\\ 
\hline
CosFace\cite{wang2018cosface} & 99.81 & 98.12 & 92.28 & 98.11 & 94.80 & 96.37 \\
ArcFace\cite{arcface} & 99.83 & 98.27 & 92.08 & 98.28 & 94.25 & 96.03 \\
MV-Softmax\cite{wang2020mis} & 99.80 & 98.28 & 92.83 & 97.95 & 93.60 & 95.20 \\
MagFace\cite{meng2021magface} & 99.83 & 98.46 & 92.87 & 98.17 & 94.51 & 95.97 \\
SCF-ArcFace\cite{9577756} & 99.82 & 98.40 & 93.16 & 98.30 & 94.74 & 96.09 \\
\cdashline{1-7}
AdaFace\cite{kim2022adaface} & 99.82 & 98.49 & 93.53 & 98.05 & 95.67 & 96.89 \\
AAFace & 99.82 & \textbf{98.56} & \textbf{93.71} & \textbf{98.24} & \textbf{95.70} & \textbf{96.93} \\
\hline \hline
\end{tabular}
  \label{Tab:FRR}
\end{table}

\section{Conclusion}
\label{Conclusion}
In this work, we proposed a multi-branch neural network that uses shared CNN feature space for two related tasks which are SB prediction and FR. The proposed architecture, not only predicts SB attributes and identifies face images simultaneously but also utilizes SB attributes as auxiliary information to improve the performance of FR. Results demonstrate that training both tasks jointly improves their performance in comparison with separate training. Moreover, experiments prove that integrating FR with SB features through our AAI module is the most effective strategy among existing integration methods.  

\small\noindent\textbf{Acknowledgment.}
This research is based upon work supported by the Office of the Director of National Intelligence (ODNI), Intelligence Advanced Research Projects Activity (IARPA), via IARPA R$\&$D Contract No. 2022-21102100001. The views and conclusions contained herein are those of the authors and should not be interpreted as necessarily representing the official policies or endorsements, either expressed
or implied, of the ODNI, IARPA, or the U.S. Government. The U.S. Government is authorized to reproduce and distribute reprints for Governmental purposes notwithstanding any copyright annotation thereon.

\bibliographystyle{IEEEbib}

 { \small\bibliography{arxive.bib}}

\begin{thebibliography}{10}

\bibitem{kim2022adaface}
Minchul Kim, Anil~K Jain, and Xiaoming Liu,
\newblock ``Adaface: Quality adaptive margin for face recognition,''
\newblock in {\em CVPR}, 2022, pp. 18750--18759.

\bibitem{arcface}
Jiankang Deng, Jia Guo, Niannan Xue, and Stefanos Zafeiriou,
\newblock ``Arcface: Additive angular margin loss for deep face recognition,''
\newblock in {\em CVPR}, 2019, pp. 4685--4694.

\bibitem{ranjan2017all}
Rajeev Ranjan, Swami Sankaranarayanan, Carlos~D Castillo, and Rama Chellappa,
\newblock ``An all-in-one convolutional neural network for face analysis,''
\newblock in {\em FG}, 2017, pp. 17--24.

\bibitem{ranjan2017hyperface}
Rajeev Ranjan, Vishal~M Patel, and Rama Chellappa,
\newblock ``Hyperface: A deep multi-task learning framework for face detection,
  landmark localization, pose estimation, and gender recognition,''
\newblock {\em IEEE Transactions on Pattern Analysis and Machine Intelligence},
  vol. 41, no. 1, pp. 121--135, 2017.

\bibitem{liu2015}
Ziwei Liu, Ping Luo, Xiaogang Wang, and Xiaoou Tang,
\newblock ``Deep learning face attributes in the wild,''
\newblock in {\em ICCV}, 2015, pp. 3730--3738.

\bibitem{moon}
Ethan~M. Rudd, Manuel G{\"u}nther, and Terrance~E. Boult,
\newblock ``Moon: A mixed objective optimization network for the recognition of
  facial attributes,''
\newblock in {\em ECCV}, 2016, pp. 19--35.

\bibitem{zhuang2018multi}
Ni~Zhuang, Yan Yan, Si~Chen, and Hanzi Wang,
\newblock ``Multi-task learning of cascaded cnn for facial attribute
  classification,''
\newblock in {\em ICPR}. IEEE, 2018, pp. 2069--2074.

\bibitem{mao2020deep}
Longbiao Mao, Yan Yan, Jing-Hao Xue, and Hanzi Wang,
\newblock ``Deep multi-task multi-label cnn for effective facial attribute
  classification,''
\newblock {\em IEEE Transactions on Affective Computing}, vol. 13, no. 2, pp.
  818--828, 2020.

\bibitem{wang2017multi}
Zhanxiong Wang, Keke He, Yanwei Fu, Rui Feng, Yu-Gang Jiang, and Xiangyang Xue,
\newblock ``Multi-task deep neural network for joint face recognition and
  facial attribute prediction,''
\newblock in {\em ICMR}, 2017, pp. 365--374.

\bibitem{a1}
Sanghyun Woo, Jongchan Park, Joon-Young Lee, and In~So Kweon,
\newblock ``Cbam: Convolutional block attention module,''
\newblock in {\em ECCV}, September 2018.

\bibitem{hu2018squeeze}
Jie Hu, Li~Shen, and Gang Sun,
\newblock ``Squeeze-and-excitation networks,''
\newblock in {\em CVPR}, 2018, pp. 7132--7141.

\bibitem{dai2021attentional}
Yimian Dai, Fabian Gieseke, Stefan Oehmcke, Yiquan Wu, and Kobus Barnard,
\newblock ``Attentional feature fusion,''
\newblock in {\em WACV}, 2021, pp. 3560--3569.

\bibitem{srivastava2015training}
Rupesh~K Srivastava, Klaus Greff, and J{\"u}rgen Schmidhuber,
\newblock ``Training very deep networks,''
\newblock {\em Advances in neural information processing systems}, vol. 28,
  2015.

\bibitem{webface260m}
Z.~Zheng et~al.,
\newblock ``Webface260m: A benchmark unveiling the power of million-scale deep
  face recognition,''
\newblock in {\em CVPR}, 2021.

\bibitem{huang2021age}
Zhizhong Huang, Junping Zhang, and Hongming Shan,
\newblock ``When age-invariant face recognition meets face age synthesis: A
  multi-task learning framework,''
\newblock in {\em CVPR}, 2021, pp. 7282--7291.

\bibitem{LFWTech}
Gary~B. Huang, Manu Ramesh, Tamara Berg, and Erik Learned-Miller,
\newblock ``Labeled faces in the wild: A database for studying face recognition
  in unconstrained environments,''
\newblock Tech. {R}ep. 07-49, University of Massachusetts, Amherst, October
  2007.

\bibitem{CFPFP}
Soumyadip Sengupta, Jun-Cheng Chen, Carlos Castillo, Vishal~M. Patel, Rama
  Chellappa, and David~W. Jacobs,
\newblock ``Frontal to profile face verification in the wild,''
\newblock in {\em WACV}, 2016, pp. 1--9.

\bibitem{CPLFWTech}
T.~Zheng and W.~Deng,
\newblock ``Cross-pose lfw: A database for studying cross-pose face recognition
  in unconstrained environments,''
\newblock Tech. {R}ep. 18-01, Beijing University of Posts and
  Telecommunications, February 2018.

\bibitem{agedb}
Stylianos Moschoglou, Athanasios Papaioannou, Christos Sagonas, Jiankang Deng,
  Irene Kotsia, and Stefanos Zafeiriou,
\newblock ``Agedb: The first manually collected, in-the-wild age database,''
\newblock in {\em CVPRW}, 2017, pp. 1997--2005.

\bibitem{whitelam2017iarpa}
Cameron Whitelam, Emma Taborsky, Austin Blanton, Brianna Maze, Jocelyn Adams,
  Tim Miller, Nathan Kalka, Anil~K Jain, James~A Duncan, Kristen Allen, et~al.,
\newblock ``Iarpa janus benchmark-b face dataset,''
\newblock in {\em CVPRW}, 2017, pp. 90--98.

\bibitem{maze2018iarpa}
Brianna Maze, Jocelyn Adams, James~A Duncan, Nathan Kalka, Tim Miller, Charles
  Otto, Anil~K Jain, W~Tyler Niggel, Janet Anderson, Jordan Cheney, et~al.,
\newblock ``Iarpa janus benchmark-c: Face dataset and protocol,''
\newblock in {\em ICB}. IEEE, 2018, pp. 158--165.

\bibitem{cao2018vggface2}
Qiong Cao, Li~Shen, Weidi Xie, Omkar~M Parkhi, and Andrew Zisserman,
\newblock ``Vggface2: A dataset for recognising faces across pose and age,''
\newblock in {\em 2018 13th IEEE international conference on automatic face \&
  gesture recognition (FG 2018)}. IEEE, 2018, pp. 67--74.

\bibitem{wang2018cosface}
Hao Wang, Yitong Wang, Zheng Zhou, Xing Ji, Dihong Gong, Jingchao Zhou, Zhifeng
  Li, and Wei Liu,
\newblock ``Cosface: Large margin cosine loss for deep face recognition,''
\newblock in {\em CVPR}, 2018, pp. 5265--5274.

\bibitem{wang2020mis}
Xiaobo Wang, Shifeng Zhang, Shuo Wang, Tianyu Fu, Hailin Shi, and Tao Mei,
\newblock ``Mis-classified vector guided softmax loss for face recognition,''
\newblock in {\em AAAI}, 2020, number~07, pp. 12241--12248.

\bibitem{meng2021magface}
Qiang Meng, Shichao Zhao, Zhida Huang, and Feng Zhou,
\newblock ``Magface: A universal representation for face recognition and
  quality assessment,''
\newblock in {\em CVPR}, 2021, pp. 14225--14234.

\bibitem{9577756}
Shen Li, Jianqing Xu, Xiaqing Xu, Pengcheng Shen, Shaoxin Li, and Bryan Hooi,
\newblock ``Spherical confidence learning for face recognition,''
\newblock in {\em CVPR}, 2021, pp. 15624--15632.

\end{thebibliography}

\end{document}